\pdfoutput=1

\documentclass[11pt]{article}
\usepackage{fix-cm}
\usepackage[preprint]{acl}

\usepackage{bm}

\usepackage{CJKutf8}
\usepackage{makecell}
\usepackage{graphicx}
\usepackage{amsmath}
\usepackage{amsfonts}
\usepackage{algorithm, algpseudocode}
\usepackage{hyperref}
\usepackage{multirow}
\usepackage{booktabs}
\usepackage{color}
\usepackage{times}
\usepackage{latexsym}
\usepackage{amssymb}
\usepackage{pifont} 

\newcommand{\xmark}{\ding{55}} 
\usepackage[T1]{fontenc}

\usepackage[utf8]{inputenc}

\usepackage{microtype}

\usepackage{inconsolata}

\usepackage[table]{xcolor} 
\usepackage{graphicx}

\usepackage{tcolorbox}
\definecolor{rq}{HTML}{1B365C}
\definecolor{rqBack}{HTML}{9ECBF7}

\title{RAL2M: Retrieval Augmented Learning-To-Match \\ Against Hallucination in Compliance-Guaranteed Service Systems}

\author{Mengze Hong$^{1}$, Di Jiang$^{1}$\thanks{Corresponding Author}, Jiangtao Wen$^{2}$, Zhiyang Su$^{3}$ \\ \textbf{Yawen Li}$^{4}$, \textbf{Yanjie Sun}$^{1}$, \textbf{Guan Wang}$^{1}$, \textbf{Chen Jason Zhang}$^{1}$\\ 
$^{1}$Hong Kong Polytechnic University \quad 
$^{2}$New York University Shanghai\\
$^{3}$Hong Kong University of Science and Technology\\
$^{4}$Beijing University of Posts and Telecommunications\\
}

\begin{document}
\maketitle
\begin{abstract}
Hallucination is a major concern in LLM-driven service systems, necessitating explicit knowledge grounding for compliance-guaranteed responses. In this paper, we introduce Retrieval-Augmented Learning-to-Match (RAL2M), a novel framework that eliminates generation hallucination by repositioning LLMs as query-response matching judges within a retrieval-based system, providing a robust alternative to purely generative approaches. To further mitigate judgment hallucination, we propose a query-adaptive latent ensemble strategy that explicitly models heterogeneous model competence and interdependencies among LLMs, deriving a calibrated consensus decision. Extensive experiments on large-scale benchmarks demonstrate that the proposed method effectively leverages the ``wisdom of the crowd'' and significantly outperforms strong baselines. Finally, we discuss best practices and promising directions for further exploiting latent representations in future work.

\end{abstract}

\section{Introduction}

Large language models (LLMs) have gradually replaced traditional search engines and dialogue systems in various service applications \cite{10.1145/3748304}. However, in compliance-critical domains such as healthcare and finance \citep{huang2024survey,ji2023survey}, regulatory requirements mandate strict human validation of service responses, raising serious concerns about the use of generative models due to the risk of hallucinations. Such issues can lead to unacceptable consequences, including financial losses, regulatory violations, and misinformation during decision-making \cite{liu2024preventing, wu2025automated}, thereby attracting significant research efforts toward hallucination mitigation.

\begin{figure}
    \centering
    \includegraphics[width=0.94\linewidth]{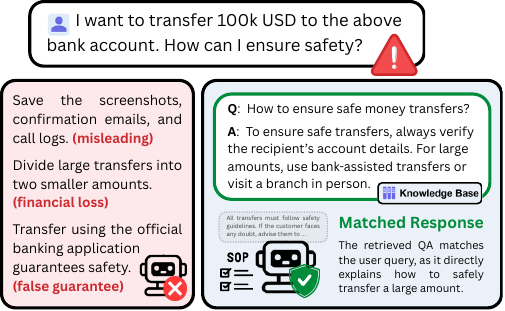}
    \vspace{-0.5em}
    \caption{System comparison under cautionary queries: generative system (left) produces hallucinated responses; retrieval system with LLM judgment (right) returns verified and appropriate response.}
    \label{fig:overview}
    \vspace{-0.9em}
\end{figure}

In practice, the bounded knowledge principle suggests drawing safe responses from a verified knowledge base. When knowledge gaps exist, fallback responses (e.g., ``I don't know'') should be used, in contrast to LLM-driven systems that often produce overconfident errors \cite{feng-etal-2024-dont, lu-etal-2025-llm}. This implies that, rather than deploying LLMs for direct user interaction, safe usage should emphasize backend system roles, such as knowledge augmentation, language understanding, and query matching (see Figure~\ref{fig:overview}). While data augmentation has been widely studied \cite{hong2025augmenting, chai2026text}, query matching remains an important yet largely underexplored component.

In this paper, we propose \textbf{Retrieval-Augmented Learning-to-Match} (RAL2M), a novel framework of LLM application in service systems that eliminates generative risks by avoiding open-ended generation while fully leveraging LLMs’ superior language understanding for accurate query-response matching. In this task, we identify a subtle failure mode: \textbf{judgment hallucination}, where existing retrieval-based models or LLM judgments frequently return poorly matched results, which severely undermine user satisfaction and system reliability. To address this, we propose a model-based ensemble technique that aggregates judgments from multiple LLMs to mitigate individual bias. Our contributions are threefold:

\begin{itemize}

\item To our knowledge, this work is the first to systematically study LLMs for query matching with a focus on hallucination mitigation, formulating the Retrieval-Augmented Learning-to-Match problem for LLM deployment with zero-generation hallucination in compliance-guaranteed service applications.

\item To mitigate judgment hallucination, we identify the LLM heterogeneity dilemma and the limitations of existing post-inference ensembles for binary outcomes, motivating a model-based Query-Adaptive Latent Ensemble approach that captures latent competence and inter-model dependencies.

\item We construct a large-scale benchmark dataset and demonstrate substantial improvements in judgment accuracy and reduced hallucination rate with the proposed method, emphasizing the importance of query-adaptive weighting and dependency structure.

\end{itemize}

\begin{table*}[t]
\centering
\resizebox{\linewidth}{!}{
\begin{tabular}{lcccc}
\toprule
\textbf{Methods} & \textbf{Query-Adaptive} & \textbf{Model Dependencies} & \textbf{Latent Features} & \textbf{Binary-Compatible} \\
\midrule
\multicolumn{5}{l}{\textit{\textbf{Static Aggregation}}} \\
Majority Voting & \xmark & \xmark & \xmark & \checkmark \\
Global Weighted Voting & \xmark & \xmark & \xmark & \checkmark \\
\midrule
\multicolumn{5}{l}{\textit{\textbf{Selection / Regeneration}}} \\
LLM-Blender \cite{jiang-etal-2023-llm} & \checkmark & \xmark & \xmark & \xmark \\
URG \cite{lv2024urg} & \checkmark & \xmark & \xmark & \xmark \\
LLM-TOPLA \cite{tekin2024llm} & \checkmark & \xmark & \xmark & \xmark \\
Agent-Forest \cite{zhangmore} & \xmark & \xmark & \xmark & \xmark \\
\midrule
\multicolumn{5}{l}{\textit{\textbf{Cascade / Routing}}} \\
FrugalGPT \cite{chenfrugalgpt} & \checkmark & \xmark & \xmark & \xmark \\
AutoMix \cite{aggarwal2024automix} & \checkmark & \xmark & \xmark & \xmark \\
EcoAssistant \cite{zhang2024ecoassistant} & \xmark & \xmark & \xmark & \xmark \\
\midrule
\multicolumn{5}{l}{\textit{\textbf{Proposed Method (Ours)}}} \\
Query-Adaptive Latent Ensemble & \checkmark & \checkmark & \checkmark & \checkmark \\
\bottomrule
\end{tabular}}
\caption{Comparison of representative post-inference ensemble methods.}
\label{tab:ensemble comparison}
\end{table*}

\section{Related Work}
\label{sec:Related Work}

\subsection{Learning-to-Match}

Learning-to-Match (L2M) is a core paradigm in information retrieval and search engines that learns to score or rank candidate responses retrieved from a knowledge base, enabling the system to return the most relevant response to a given query~\cite{mitra2017learning, xu2024learning}. It operates under a strict compliance constraint: the system either returns an exact, human-verified answer from the knowledge base or explicitly refuses when none is sufficiently relevant, typically indicated by a very low matching score or an unmatched decision. This contrasts with Retrieval-Augmented Generation (RAG), where retrieved documents primarily serve as contextual input for an LLM to synthesize new responses~\cite{fan2024survey, li2025matching}. L2M is often formulated as a cost-efficient binary decision problem, in which the system determines whether the top-ranked candidate adequately matches the user query~\cite{shao2023understanding}.

Early L2M models relied on dual encoders~\cite{hu2014convolutional} or interaction-based matching~\cite{wan2016match}, primarily due to their efficiency and suitability for large-scale retrieval, whereas modern approaches utilize pre-trained bi-encoders or cross-encoders~\cite{reimers2019sentence} to better capture semantic similarity. Despite these advances, existing L2M systems exhibit key limitations: they struggle to adapt flexibly to diverse domain-specific knowledge and fail to capture intent-level distinctions across topics or expressions \cite{li-gaussier-2024-domain}. These shortcomings motivate RAL2M, which replaces the traditional scalar similarity score with LLM judgments that have proven capability in relevance generation and ranking \cite{sun-etal-2023-chatgpt}, conditioned on the semantic understanding of the user query, response candidates, and supporting documents to achieve higher precision in identifying suitable matches.

\subsection{LLM Ensemble}

Recognizing the persistent bias and hallucination of generative models, LLM ensembling, which leverages multiple models to capitalize on their individual strengths, has proven effective in improving downstream performance \cite{chen2025harnessing}. For example, \citet{yang2023one} proposed an ensemble pipeline integrating state-of-the-art LLMs to enhance accuracy and reliability across medical QA datasets, addressing domain-specific challenges where precision is crucial. Similarly, \citet{niimi2025simple} introduced a simple ensemble strategy for sentiment analysis, and \citet{abburi2023generative} applied ensembling to generative text classification, both improving stability and reproducibility in LLM outputs. Unlike typical multi-agent systems, which involve interactive collaboration or debate during inference \cite{chan2024chateval}, post-inference LLM ensembles operate independently without inter-agent communication. This makes them particularly suitable for service systems, where multiple LLMs can operate in parallel during online inference to reduce system latency \cite{rodionov2025hogwild}.

Despite these advantages, existing ensemble strategies face three major limitations: 1) binary decisions carry minimal information, making aggregation challenging \cite{galar2011overview}; 2) fixed voting or weighting schemes generalize poorly across diverse queries and domains \cite{ai2025beyond}; and 3) cascade-based methods fail because a single binary output provides insufficient signal to guide subsequent routing \cite{soiffer2025semantic}. Addressing these challenges requires incorporating external features. In this paper, we propose leveraging query-conditional contextual signals and latent model-centric information, characterized by LLM topical competencies and inter-model dependencies, clearly distinguishing the proposed method from existing approaches (see Table \ref{tab:ensemble comparison}).

\section{Preliminaries}

\subsection{Problem Formulation}

Given a structured knowledge base of canonical question-answer pairs \(D_1 = \{(q_i, a_i)\}_{i=1}^N\) and an unstructured evidence base of textual documents \(D_2\), the RAL2M problem aims to determine whether a retrieved candidate response correctly matches a user query \(q\) by estimating a binary relevance probability:
\[
p(y=1 \mid q, R^q_{D_2}, q_i, a_i) \in [0,1],
\]
where \(y=1\) indicates that \((q_i, a_i)\) matches the user's query and is a valid response, supported by evidence \(R^q_{D_2} \subseteq D_2\). The training objective is to minimize a loss (e.g., binary cross-entropy) over labeled examples 
$(q, R^q_{D_2}, q_i, a_i, y) \sim \mathcal{D}$:
\begin{equation*}
p^* = \operatorname*{arg\,min}_{p}
\mathbb{E}_{\mathcal{D}}
\Big[
\ell\big(p(y \mid q, R^q_{D_2}, q_i, a_i), y\big)
\Big].
\end{equation*}

In inference, the retrieved candidate is returned only if the estimated probability exceeds a threshold; otherwise, a safe fallback is returned to guarantee zero generation hallucination.

\subsection{The LLM Heterogeneity Dilemma}

Although LLM ensembling is expected to mitigate individual biases via the ``wisdom of the crowd'' effect, its practical benefits are often limited \cite{arabzadeh2025benchmarking}. Pre-training on large, internet-sourced corpora with substantial overlap induces heterogeneous dependencies across models, resulting in diverse correlation patterns, varying inter-model agreement, and task-specific strengths and weaknesses \cite{abels2025wisdom, pmlr-v267-kim25e}. Such heterogeneity presents a fundamental challenge for ensemble design: naive voting or selection-based approaches often yield marginal improvement over the strongest single model \cite{hong-etal-2025-dial, lefort2024examiningindependenceensemblesentiment} and can even amplify systematic errors when a strong model is outvoted by multiple weaker but highly correlated models \cite{chen2024are, hong-etal-2025-qualbench}. This challenge is especially pronounced in binary judgment ensembles with more diverse patterns of correlation and inter-annotator agreement (see Table \ref{tab:correlation}), necessitating explicit modeling of latent dependencies.

\section{Proposed Methods}

\begin{figure}
    \centering
    \includegraphics[width=0.98\linewidth]{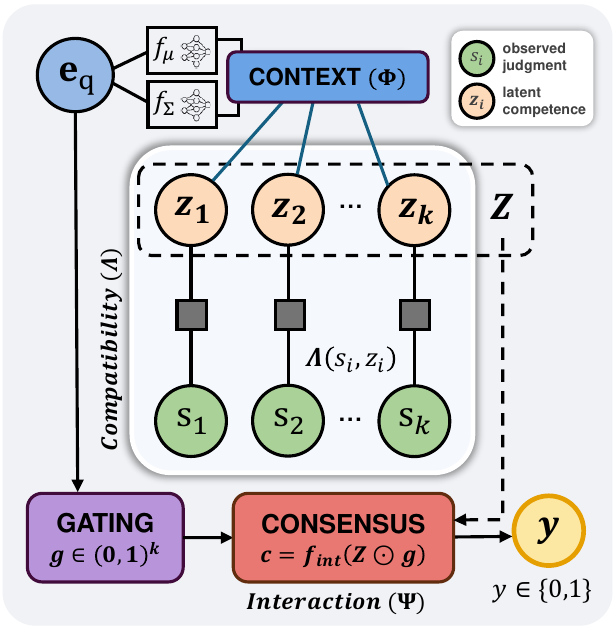}
    \caption{Model architecture of query-adaptive latent ensemble with query embedding $\mathbf{e}_q$ and prediction $y$.}
    \label{fig:model}
\end{figure}

To mitigate judgment hallucination and improve query matching performance, we propose \textbf{Query-Adaptive Latent Ensemble}, a model-based framework that explicitly models latent dependencies among $k$ heterogeneous LLM judges in a post-inference ensemble. Given user query $q$, a retrieved candidate response $(q',a')$, and binary judgments $\mathbf{s} = (s_1, \dots, s_k)$ from $k$ LLMs:
\begin{equation}
\begin{aligned}
s_i(q,(q',a'))=LLM_i(q,(q',a'), R^q_{D_2}),
\end{aligned}
\label{ngram}
\end{equation}
the goal is to infer a calibrated binary relevance decision $y \in \{0,1\}$ indicating whether the retrieved QA pair correctly matches the user query. If $y=0$, a safe fallback response is returned.

\subsection{Model Architecture}

Intuitively, the key to combining binary judgments does not lie in the judgment content itself, which carries very limited information, but in identifying the model most competent for a given query, leveraging its capability to provide the most reliable judgment \cite{masoudnia2014mixture, zhou2022mixture}. As shown in Figure \ref{fig:model}, the proposed model introduces a continuous latent variable $\mathbf{Z} = (\mathbf{z}_1, \ldots, \mathbf{z}_k) \in \mathbb{R}^k$, where each $\mathbf{z}_i$ encodes the query-dependent latent competence of the $i$-th LLM. Non-linear dependencies among LLMs, along with uncertainties and consensus formation, are jointly captured by $\mathbf{Z}$ within a graphical model. Formally, we define a conditional Gibbs distribution over the joint space $(y, \mathbf{Z})$ given observed judgments $\mathbf{s}$ and query embedding $\mathbf{e}_q$:
\begin{equation}
P(y, \mathbf{Z} \mid \mathbf{s}, \mathbf{e}_q) = \frac{1}{\mathcal{Z}(\mathbf{s}, \mathbf{e}_q)}
\exp\left( - E(y, \mathbf{Z}; \mathbf{s}, \mathbf{e}_q) \right),
\end{equation}
where $\mathcal{Z}$ is the partition function normalizing the probability. This Gibbs formulation allows joint prediction of $y$ and estimation of $\mathbf{Z}$ by minimizing a global energy function $E$, finding the equilibrium where historical reliability, current votes, and group consensus are most consistent.

To effectively capture query-dependent competence, model consistency, and decision-level interactions, we design the energy function to decompose additively into three potentials, each regularizing a distinct aspect of the inference:
\begin{equation}\small
E(y, \mathbf{Z}; \mathbf{s}, \mathbf{e}_q)
=
\underbrace{\Phi(\mathbf{Z}; \mathbf{e}_q)}_{\text{Context}}
+
\underbrace{\Lambda(\mathbf{s}, \mathbf{Z})}_{\text{Compatibility}}
+
\underbrace{\Psi(y, \mathbf{Z}; \mathbf{e}_q)}_{\text{Interaction}}.
\end{equation}

\subsection{Model Parameterization}
\paragraph{Query-conditioned Context Potential $\Phi(\mathbf{Z}; \mathbf{e}_q)$.}
With query embedding $\mathbf{e}_q \in \mathbb{R}^{d}$ obtained from a dense encoder with rich semantic information, we define a diagonal Gaussian distribution over the latent competence variables $\mathbf{Z}$:
\begin{equation}
p(\mathbf{Z} \mid \mathbf{e}_q) = \mathcal{N}\left(\mathbf{Z} \mid \boldsymbol{\mu}(\mathbf{e}_q), \boldsymbol{\sigma}(\mathbf{e}_q)\right),
\end{equation}
parameterized by two weight-normalized neural networks: $\boldsymbol{\mu}(\mathbf{e}_q) = f_\mu(\mathbf{e}_q)$, capturing expected judge competence based on patterns in the query embedding, and $\boldsymbol{\sigma}(\mathbf{e}_q) = \mathrm{softplus}(f_\Sigma(\mathbf{e}_q)) + \epsilon$, which quantifies uncertainty in those expectations. 

The amortized networks $f_\mu$ and $f_\Sigma$ are trained end-to-end on the labeled data to produce an adaptive, query-dependent distribution that emphasizes competent judges, increases uncertainty for ambiguous queries, and keeps the posterior aligned with semantically similar queries. To enforce this, we define the corresponding energy term as the analytic KL divergence to a standard normal prior:
\begin{equation}\small
\Phi(\mathbf{Z}; \mathbf{e}_q) = \frac{1}{2} \sum_{i=1}^{k} \left( \boldsymbol{\mu}_i(\mathbf{e}_q)^2 + \boldsymbol{\sigma}_i^2(\mathbf{e}_q) - \log \boldsymbol{\sigma}_i^2(\mathbf{e}_q) - 1 \right).
\end{equation}

\paragraph{Compatibility Potential $\Lambda(\mathbf{s}, \mathbf{Z})$.}
The compatibility potential aligns each binary judgment $s_i$ with its latent competence $\mathbf{z}_i$ via a linear function:
\begin{equation}
\Lambda(\mathbf{s}, \mathbf{Z}) = \sum_{i=1}^{k} \left( \theta_{\phi,i} \mathbf{z}_i + W_{\phi,i} s_i \mathbf{z}_i \right),
\end{equation}
where $\theta_{\phi,i}$ sets a base expectation for competence and $W_{\phi,i}$ adjusts the influence of observed judgments. This term encourages consistency between judgments and latent competence while allowing flexibility to accommodate disagreement or uncertainty in the sparse binary signals.

\paragraph{Interaction Potential $\Psi(y, \mathbf{Z}; \mathbf{e}_q)$.}
To capture higher-order dependencies among judges and drive consensus toward a global label $y$, we introduce a gated non-linear interaction:
\begin{equation}
\mathbf{g} = \sigma(f_g(\mathbf{e}_q)) \in (0,1)^k,
\end{equation}
which selectively emphasizes relevant judges (higher $g_i$ weights judge $i$ more). The modulated latents are aggregated by a shallow MLP into a scalar consensus score $c = f_{\text{int}}(\mathbf{Z} \odot \mathbf{g})$,
learned to be positive for hallucinated patterns ($y=0$) and negative for correct matches ($y=1$).

The interaction potential is then defined as
\begin{equation}
\Psi(y, \mathbf{Z}; \mathbf{e}_q) = \theta_{\lambda,y} \cdot c,
\end{equation}
with $\theta_{\lambda,0} < 0$ and $\theta_{\lambda,1} > 0$. A positive consensus score $c$ lowers the energy for $y=0$, while a negative $c$ lowers the energy for $y=1$. This asymmetric bilinear form rewards labels aligned with the collective gated judgment, enabling query-adaptive decisions without requiring explicit pairwise factors. By parameterizing $\Psi$ through the shared consensus score $c$, the latent variables $\mathbf{z}_i$ are implicitly coupled, allowing the model to learn inter-judge correlations without the quadratic computational cost of a fully connected graph.

\subsection{Posterior Approximation}

Exact inference of the joint posterior $P(y, \mathbf{Z} \mid \mathbf{s}, \mathbf{e}_q)$ is intractable due to the continuous latent space and non-linear neural potentials \cite{Kingma2014}. Instead of explicit marginalization, we estimate the latent competence $\mathbf{Z}$ by identifying the mode of the energy function via damped fixed-point iterations \cite{bengio2015early}. The update for the mean $\boldsymbol{\mu}_i$ integrates the contributions of the three energy potentials:

\begin{equation}
\label{eq:update}
\boldsymbol{\mu}_i^{(t+1)} = \dfrac{\mathcal{C}_i + \mathcal{L}_i + \mathcal{I}_i}{1 / \boldsymbol{\sigma}_i^{2}(\mathbf{e}_q) + 1},
\end{equation}
where the terms are defined as:
\begin{itemize}

\item \textbf{Context term ($\mathcal{C}_i$):} $\boldsymbol{\mu}_i(\mathbf{e}_q) / \boldsymbol{\sigma}_i^{2}(\mathbf{e}_q)$, representing the baseline competence of judge $i$ for the given query.

\item \textbf{Compatibility term ($\mathcal{L}_i$):} $\theta_{\phi,i} + W_{\phi,i} s_i$, adjusting the latent competence based on the observed judgment.

\item \textbf{Interaction term ($\mathcal{I}_i$):} $g_i \cdot\mathbb{E}_{y \sim P(y \mid \mathbf{Z})}[\theta_{\lambda,y}]$, capturing the alignment with the ensemble consensus under query-dependent gating.
\end{itemize}

To ensure numerical stability, we apply momentum damping \cite{lucas2019} to the update: $\boldsymbol{\mu}^{(t+1)} \leftarrow \alpha \cdot \boldsymbol{\mu}_{\text{new}}^{(t+1)} + (1 - \alpha) \cdot \boldsymbol{\mu}^{(t)}$, where $\alpha > 0.5$. The posterior variances are fixed analytically as $v_i = 1 / (1/\boldsymbol{\sigma}_i^{2}(\mathbf{e}_q) + 1)$.

\begin{algorithm}[!t]\small
\caption{Online Inference}
\begin{algorithmic}[1]
\State \textbf{Input:} query embedding $\mathbf{e}_q$, LLM judgments $\mathbf{s}$
\State Compute $\boldsymbol{\mu}(\mathbf{e}_q), \boldsymbol{\sigma}^2(\mathbf{e}_q), \mathbf{g} = \sigma(f_g(\mathbf{e}_q))$
\State Posterior variances: $\mathbf{v} \gets 1 / (1/\boldsymbol{\sigma}^2(\mathbf{e}_q)+1)$
\State Refine $\boldsymbol{\mu}$ via damped fixed-point updates (Eq.~\ref{eq:update}) using $\mathbf{s}$
\State Draw $M$ samples: $\mathbf{z}^{(m)} \sim \mathcal{N}(\boldsymbol{\mu}, \text{diag}(\mathbf{v}))$
\State $\hat{p} \gets \frac{1}{M}\sum_{m=1}^M \sigma(f_{\text{int}}(\mathbf{g} \odot \mathbf{z}^{(m)}) \cdot \theta_\lambda)$; $y \gets \mathbb{I}[\hat{p} > 0.5]$
\State \textbf{Output:} $y$
\end{algorithmic}
\label{alg:inference}
\end{algorithm}

\subsection{Training and Inference}

\paragraph{Offline Training.}
The marginal probability $\hat{p} = P(y=1 \mid \mathbf{s}, \mathbf{e}_q)$ is approximated via Monte Carlo sampling from the inferred latent distribution:
\vspace{-2em}
\begin{equation}
\begin{aligned}
&\mathbf{z}^{(m)} \sim \mathcal{N}(\boldsymbol{\mu}, \text{diag}(\mathbf{v})), \quad\\&
\hat{p} = \frac{1}{M} \sum_{m=1}^M \sigma\!\left( f_{\text{int}}(\mathbf{g} \odot \mathbf{z}^{(m)}) \cdot \theta_\lambda \right).  
\end{aligned}
\end{equation}

The model is optimized using focal binary cross-entropy loss \cite{mukhoti2020calibrating} to focus on hard cases, combined with label smoothing to prevent overconfidence. An annealed KL divergence regularizes the inferred $\mathbf{Z}$ toward the query-conditioned Gaussian prior $\Phi(\mathbf{Z}; \mathbf{e}_q)$.

\paragraph{Online Inference.}
With frozen parameters, given a new query embedding $\mathbf{e}_q$ and LLM judgments $\mathbf{s}$, the final judgment is obtained as in Algorithm \ref{alg:inference}.

\section{Experimental Setup}
\subsection{Dataset}

We construct a large-scale multi-domain benchmark by extending five public QA collections \cite{friel2024ragbench}, resulting in a verified knowledge base of 10,020 QA pairs and 47,975 supporting document chunks. For each canonical QA pair, we simulate two types of user queries:

\begin{itemize}
    \item \textbf{Aligned queries} (positive): semantically and intentionally equivalent to $q_i$. The ground-truth response is $a_i$.
    \item \textbf{Misaligned queries} (negative): topically related but intentionally misaligned to $q_i$. The ground-truth is a safe fallback response.
\end{itemize}

\noindent Both types of queries are generated via intention-enhanced similar-question generation~\cite{hong2025augmenting} under strict and relaxed semantic constraints, ensuring sufficient distinctiveness to capture diverse user query expressions (see Table \ref{tab:dataset_stats}). 

To ensure data quality, two LLM judges first classify each query for alignment with its source QA, followed by cleaning and filtering by five human annotators, yielding 82,606 simulated queries. The dataset is balanced, with an equal number of positive and negative samples. Negative samples are particularly valuable, representing topically relevant queries outside the knowledge boundary, which forces models to learn precise intent rather than rely on superficial similarity. For each query, the top-1 candidate response and its supporting document chunk are retrieved, and the dataset is split 70\% for training and 30\% for testing\footnote{Data and code are publicly available at \href{https://anonymous.4open.science/r/RAL2M-2026/}{GitHub Repo}. Examples of source question and query are shown in Table \ref{tab:positive_negative_query_example}.}.

\begin{table}[!t]
\centering
\small
\resizebox{\columnwidth}{!}{
\begin{tabular}{lccc}
\toprule
\textbf{Dataset} & \textbf{Questions} & \textbf{Queries} \\ 
\midrule
HotpotQA \cite{yang-etal-2018-hotpotqa} & 2,655 & 21,743 \\
MS MARCO \cite{nguyen2016ms}            & 2,550 & 19,786 \\
CovidQA \cite{moller-etal-2020-covid}  & 1,707 & 14,961 \\
ExpertQA \cite{malaviya-etal-2024-expertqa} & 1,766 & 14,089 \\
HAGRID \cite{kamalloo2023hagrid}        & 1,342 & 12,027 \\
\midrule
\textbf{Total} & \textbf{10,020} & \textbf{82,606} \\
\bottomrule
\addlinespace[1ex]
\textbf{Distinct$_N$ (\%)}: & $D_5$: 89.2 & $D_7$: 73.6 \\
\bottomrule
\end{tabular}}
\caption{Overview of the RAL2M dataset.}
\label{tab:dataset_stats}
\vspace{-1em}
\end{table}

\subsection{Evaluation Metrics}

We evaluate the baseline and proposed method using \textbf{Accuracy}, measuring the percentage of samples with correct relevance judgment. Positive samples are evaluated based on alignment between retrieved response and ground truth, while all negative samples have no matched response (i.e., $y=0$). We further quantify judgment hallucination rate as cases where false responses are incorrectly accepted, measured by the \textbf{false positive rate} \cite{koyejo2014consistent}, where higher values indicate a bias toward fabricated matches. In addition, we report \textbf{Precision} and \textbf{F1-score} for the positive class to provide a comprehensive evaluation.

\subsection{Baselines}

We compare representative baselines across three stages of a retrieval-based service system: data-level, inference-level, and decision-level. At the data level, keyphrase \cite{viswanathan-etal-2024-large} and intent \cite{hong-etal-2025-dial} expansions concatenate query embeddings with summarized information to assess whether enriched representations improve response matching. At the inference level, we adopt LLM debate following \cite{10.1145/3589334.3645381}, where models generate explainable judgments with justifications, and consensus is derived via an additional LLM. At the decision level, we compare majority voting, weighted voting, and a neural aggregation baseline that learns query-adaptive weights without explicitly capturing interactions, serving as a strong ablation to evaluate the effect of inter-model dependencies.

\subsection{Model Implementation}

We selected five open-source LLMs\footnote{Models include \href{https://huggingface.co/Qwen/Qwen2.5-7B}{Qwen2.5-7B}, \href{https://huggingface.co/mistralai/Mistral-7B-Instruct-v0.2}{Mistral-7B-v0.2}, \href{https://huggingface.co/meta-llama/Llama-3.1-8B}{Llama-3.1-8b}, \href{https://huggingface.co/google/gemma-2-9b}{Gemma-2-9b}, \href{https://huggingface.co/zai-org/glm-4-9b-chat}{GLM-4-9b}, and GPT-5 for ablation.}, chosen for their strong capabilities and lightweight deployment, covering a range of model sizes \cite{tan2025judgebench}. The \texttt{bge-large-en-v1.5} is used as the dense encoder for query retrieval. Experiments were conducted on a single NVIDIA L20 GPU with 48GB of memory, enabling batch inference and parallel processing. The judgment prompt was tuned using PromptWizard \cite{agarwal-etal-2025-promptwizard} to ensure high-quality instruction following and consistent outputs. All LLMs were initialized with a temperature of 0 to ensure deterministic behavior. Hyperparameter tuning was performed to identify optimal configurations for key settings, including hidden dimensions, dropout rate, learning rate, number of Monte Carlo samples, and the number of variational inference iterations. Each ensemble model was run with five different seeds and reported the average performance for robustness\footnote{More details are provided in Appendix \ref{appendix:profile and prompt}.}.

\begin{figure*}[!t]
    \centering
    \includegraphics[width=0.9\linewidth]{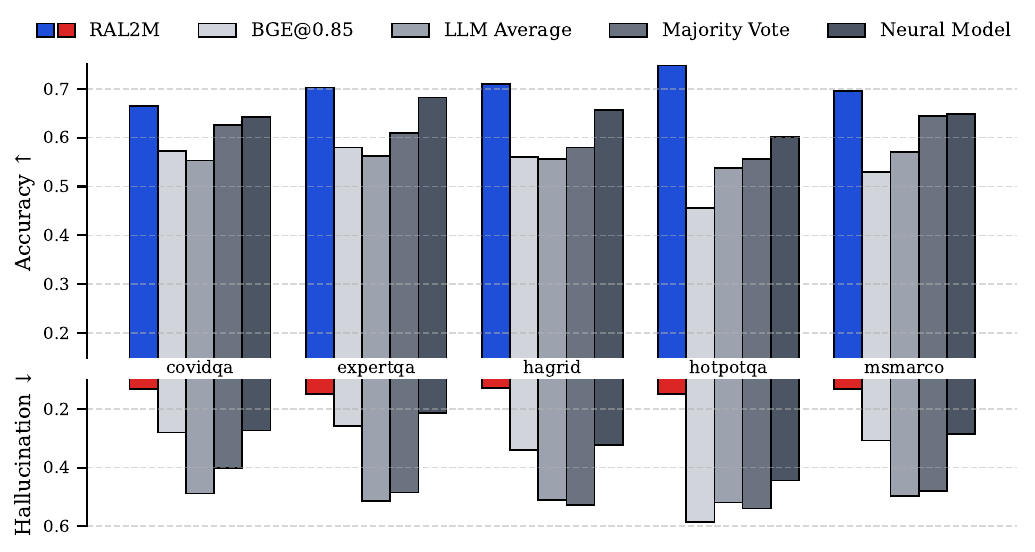}
    \vspace{-1em}
    \caption{Comparison of accuracy and hallucination rate across datasets.}
    \label{fig:main results}
    \vspace{-1em}

\end{figure*}

\section{Results and Discussions}

\subsection{Main Results}

Table \ref{tab:main results} and Figure \ref{fig:main results} present the performance comparison. Traditional threshold-based retrieval with data augmentation performs poorly in this task, highlighting the difficulty of reliably rejecting mismatched responses based on a calibrated decision threshold. Keyphrase augmentation yields modest gains by enriching semantic representations, whereas intent augmentation, which relies on overly concise intent expressions, biases the system toward accepting most candidates, resulting in extremely high hallucination rates. Single LLM judgments remain unreliable for accurately identifying matched responses, despite the high efficiency of processing approximately 23 queries per second. LLM debate nearly eliminates hallucination (1\%) but yields extremely low F1 scores due to excessive rejection, limiting its practical applicability.

\begin{table}[!t]
\centering
\resizebox{\columnwidth}{!}{
\begin{tabular}{lcccc}
\toprule
\textbf{Method} & \textbf{Acc (\%)} & \textbf{Hallu (\%)} & \textbf{Precision} & \textbf{F1} \\ \midrule
\multicolumn{5}{c}{\textit{Retrieval with Data Augmentation}} \\ \midrule
BGE@$0.85$ & 53.1 & 38.3 & 0.449 & 0.432 \\
BGE+Key & 58.3 & 34.0 & 0.510 & 0.494 \\ 
BGE+Intent & 49.5  &    98.8   &  0.496  & 0.560  \\\midrule
\multicolumn{5}{c}{\textit{LLM Inference}} \\ \midrule
LLM Average & 55.5 & 50.6 & 0.486 & 0.552 \\
GPT-5  &  55.7  &  61.3 & 0.489 & 0.602 \\
LLM Debate &  59.7   &    1.1  &   0.838 & 0.135 \\ \midrule
\multicolumn{5}{c}{\textit{LLM Ensemble}} \\ \midrule
Majority Vote & 60.2 & 49.2 & 0.526 & 0.611 \\
Weighted Vote & 59.6 & 41.1 & 0.525 & 0.563 \\
Neural Model & 64.2 & 32.5 & 0.580 & 0.588 \\
\textbf{Latent Model (ours)} & \textbf{70.7} & \textbf{13.9} & \textbf{0.730} & \textbf{0.634} \\ \bottomrule
\end{tabular}}
\caption{Comparison of baseline methods and the proposed latent ensemble for retrieval judgment.}
\label{tab:main results}
\vspace{-1em}
\end{table}

LLM ensemble approaches generally outperform single models, demonstrating the benefit of aggregating multiple judgments. Majority and weighted voting yield moderate gains, improving accuracy by 5--7\% while slightly lowering hallucination. The discriminative neural model further boosts accuracy and reduces hallucination, highlighting the value of query-adaptive weighting. However, these methods still exhibit over 30\% hallucination rate, which undermines system reliability and user satisfaction. Our proposed method achieves the highest accuracy and lowest hallucination, along with the highest precision and F1, underscoring the critical importance of modeling both query context and inter-model dependencies in post-inference ensembles. The 70\% accuracy highlights the task complexity and motivates future research to develop more tailored solutions \footnote{Model convergence and additional results on the Chinese dataset are provided in Appendix \ref{appendix:convergence} and \ref{appendix:chinese}.}.

\subsection{Effect of Retrieval Augmentation}

Incorporating retrieved document chunks generally enhances judgment performance, as evidenced by the increased accuracy of most LLMs in Table~\ref{tab:llm_accuracy}. However, augmented input can slightly distract models, especially when additional content increases sequence length or introduces peripheral information. This underscores the importance of considering both document accessibility and relevance in practical deployment.

\begin{table}[!t]
\centering
\resizebox{\columnwidth}{!}{
\begin{tabular}{lccccc}
\toprule
 & Qwen & Llama & Mistral & Gemma & ChatGLM \\
\midrule
Original & 63.9 & 61.5 & 42.9 & 58.6 & 48.9  \\
Augmented & 61.1 & 63.6 & 43.3 & 60.9 & 51.5 \\
\bottomrule
\end{tabular}}
\caption{Accuracy (\%) of single LLM inference.}
\label{tab:llm_accuracy}
\vspace{-0.5em}
\end{table}

\begin{table}[!t]
\centering
\resizebox{\columnwidth}{!}{
\begin{tabular}{lccccc}
\toprule
 & \textbf{Qwen} & \textbf{Gemma} & \textbf{Llama} & \textbf{ChatGLM} & \textbf{Mistral} \\
\midrule
\textbf{Qwen}     
    & \cellcolor{blue!20}1.000 & \cellcolor{blue!10}0.514 & \cellcolor{blue!10}0.440 & \cellcolor{blue!10}0.301 & \cellcolor{blue!10}0.041 \\
\textbf{Gemma}    
    & \cellcolor{green!10}0.464 & \cellcolor{blue!20}1.000 & \cellcolor{blue!10}0.417 & \cellcolor{blue!10}0.441 & \cellcolor{blue!10}0.068 \\
\textbf{Llama}    
    & \cellcolor{green!10}0.440 & \cellcolor{green!10}0.384 & \cellcolor{blue!20}1.000 & \cellcolor{blue!10}0.297 & \cellcolor{blue!10}0.043 \\
\textbf{ChatGLM}  
    & \cellcolor{green!10}0.170 & \cellcolor{green!10}0.348 & \cellcolor{green!10}0.174 & \cellcolor{blue!20}1.000 & \cellcolor{blue!10}0.097 \\
\textbf{Mistral}  
    & \cellcolor{green!10}0.004 & \cellcolor{green!10}0.009 & \cellcolor{green!10}0.004 & \cellcolor{green!10}0.027 & \cellcolor{blue!20}1.000 \\
\bottomrule
\end{tabular}}
\caption{Inter-dependency among five LLM judges. Upper triangle: Pearson correlation ($r$); Lower triangle: Inter-annotator agreement with Cohen's Kappa ($\kappa$).}
\label{tab:correlation}
\vspace{-0.5em}

\end{table}

\subsection{Judgment Dependencies and Distributions}

To understand the rationale behind performance gains in the proposed method, we analyze inter-model dependencies and judgment distributions. Table \ref{tab:correlation} reveals a heterogeneous dependency structure: under identical inference settings, Mistral is largely uncorrelated with other models, whereas Gemma and Qwen exhibit strong correlation. This variability challenges simple voting-based ensembles, which assume equal model reliability and cannot account for redundancy or complementary behaviors. In contrast, query-adaptive ensembles dynamically weight each model based on the query context, contributing to the superior performance of the neural model and our latent ensemble.

\begin{figure}[!t]
    \centering
    \includegraphics[width=1\linewidth]{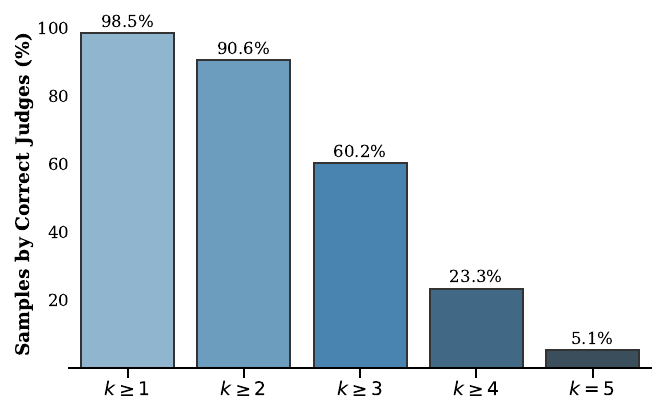}
    \vspace{-2em}
    \caption{Distribution of the number of correct judges.}
    \label{fig:correct judge}
\end{figure}

From Figure \ref{fig:correct judge}, a surprising result shows that 98.5\% of samples have at least one correct judge, and 60.2\% of samples have at least three models in agreement, which aligns with majority vote accuracy. This reveals that to boost performance beyond majority voting, the core challenge is to recover the 40\% of samples where minority judgments are more accurate. Our proposed method addressed this by explicitly modeling inter-LLM dependencies and query-adaptive competence, enabling it to leverage minority but reliable judgments to improve overall accuracy while avoiding over-reliance on correlated or dominant models.

\subsection{Ablation Studies}

Single-LLM inference shows that judgment quality is only weakly correlated with model size; for example, GPT-5 does not outperform smaller models. Ablation studies highlight two key drivers of ensemble performance: the scale of training data and the number of models included.

\subsubsection{Training Data Size}

\begin{table}[!t]
\centering
\small
\begin{tabular}{ccc}
\toprule
Training Data (\%) & Acc (\%) & Hallu (\%) \\
\midrule
10 & 56.5 & 47.8 \\
30 & 59.5 & 40.3 \\
50 & 67.6 & 27.2 \\
70 & 68.8 & 22.2 \\
90 & 69.1 & 21.4 \\
\bottomrule
\end{tabular}
\caption{Performance of ensemble model with varying training data size.}
\label{tab:ablation training size}
\vspace{-0.5em}
\end{table}

We split the training data into 10 balanced subsets and progressively increase the data size for training the latent model. As shown in Table \ref{tab:ablation training size}, performance improves consistently with more data: using only 10\% (5,756 queries) results in lower accuracy and higher hallucination than majority voting, while scaling from 10\% to 90\% yields a 12.6\% accuracy gain and a 26.4\% hallucination reduction. This reflects the fact that in low-data regimes, latent features cannot be reliably learned. However, the proposed method surpasses all baselines at 50\% training data, indicating that moderate supervision is sufficient for effective hallucination mitigation, a favorable scalability in practical deployment. Since the benchmark spans multiple domains, applying the method to a single-domain use case would require significantly less data.

\begin{table}[!t]
\centering
\resizebox{\columnwidth}{!}{
\begin{tabular}{c|cc|cc}
\toprule
\multirow{2}{*}{Models} & \multicolumn{2}{c|}{Ours} & \multicolumn{2}{c}{Majority Voting} \\
\cline{2-5}
 & Acc (\%) & Hallu (\%) & Acc (\%) & Hallu (\%) \\
\midrule
2 & 65.8 & 28.9 & 58.6 & 42.0 \\
3 & 66.2 & 27.3 & 48.4 & 87.2 \\
4 & 69.1 & 26.4 & 60.9 & 41.8 \\
5 & 70.9 & 13.9 & 60.2 & 49.2 \\
\bottomrule
\end{tabular}}
\caption{Comparison of proposed and majority voting ensemble approaches with ensemble size.}
\label{tab:ablation model count}
\vspace{-1em}
\end{table}

\subsubsection{Are More Models All We Need?}
Table~\ref{tab:ablation model count} compares ensemble performance as models are added sequentially at random. While increasing the ensemble size generally benefits both methods, majority voting exhibits unstable behavior, with severe performance degradation and extremely high hallucination when model disagreements intensify. In contrast, the proposed method consistently improves as more models are incorporated, achieving higher accuracy and substantially lower hallucination rates. Notably, even with fewer models, our method still outperforms majority voting, showing that capturing meaningful latent features is more effective than merely increasing ensemble size.

\section{Conclusion}
\label{sec:Conclusion}

This paper introduces Retrieval-Augmented Learning-to-Match, a novel framework that eliminates generation hallucination in compliance-guaranteed service systems by repositioning locally-deployed lightweight LLMs as a query-response matching utility. To mitigate judgment hallucination in returning poorly matched responses, we propose a query-adaptive latent ensemble that models heterogeneous competencies and inter-model dependencies among LLMs, substantially improving matching accuracy and reducing hallucination with high inference efficiency. This promising performance lays the foundation for future research to further exploit latent variables for more flexible LLM ensembles, particularly in tasks with limited observed signals (e.g., binary outcomes) but rich latent information.

\section{Limitations}
\label{sec:Limitation}

While Retrieval-Augmented Learning-to-Match demonstrates promising performance in mitigating both types of hallucinations in the service system, addressing the following limitations will be crucial for further refining and broadening the applicability of the framework.

\begin{enumerate}
    \item \textbf{Reliance on Knowledge Base Completeness and Accuracy:} RAL2M's fundamental strength lies in its ability to retrieve and validate answers against a pre-existing knowledge base ($D_1$). Consequently, its performance is directly constrained by the completeness and accuracy of this knowledge base. If $D_1$ lacks an answer to a novel or highly specific query, RAL2M will correctly defer to a fallback response. However, this means it cannot learn or synthesize information not already present, unlike generative models. In domains requiring answers to constantly evolving or emergent topics not yet codified in $D_1$, RAL2M would require frequent updates to its knowledge base, which can be resource-intensive.

    \item \textbf{Computational Cost of LLM Ensemble:} Although the proposed latent model provides robust hallucination mitigation, it requires multiple LLM inferences per user query, which can be more computationally expensive than a single generative LLM call. For high-stakes domains where user satisfaction and information integrity are crucial, we consider this computational investment justified, especially given our ablation results, which show that even with limited training data and a smaller ensemble, the model still outperforms existing methods. Nonetheless, future research on enhancing single-LLM performance and developing more efficient strategies could further improve practicality and create more impact.
\end{enumerate}

These limitations point to several directions for future research, including optimizing large-scale knowledge base management, developing more efficient LLM inference strategies, and refining dependency modeling to better accommodate diverse and evolving LLM architectures.

\bibliography{custom}

\appendix
\newpage

\section{Implementation Details}
\label{appendix:profile and prompt}

\subsection{Metrics Calculation}

Let $y_i \in \{0,1\}$ denote the ground-truth label of sample $i$, where $1$ indicates a matched query-response pair, and let $\hat{y}_i \in \{0,1\}$ denote a judgment from either a single or an ensemble model. Based on these labels, true positives (TP), false positives (FP), true negatives (TN), and false negatives (FN) are defined under the standard discrimination evaluation setting for binary classification \cite{lever2016classification}. The reported metrics are calculated as follows:

\begin{itemize}
    \item \textbf{Accuracy} measures the overall correctness of the judgment decisions and is defined as
    \begin{equation}
    \mathrm{Accuracy} = \frac{\mathrm{TP} + \mathrm{TN}}{\mathrm{TP} + \mathrm{TN} + \mathrm{FP} + \mathrm{FN}} .
    \end{equation}

    \item \textbf{Hallucination Rate} quantifies the likelihood of accepting an incorrect response and is defined as
    \begin{equation}
    \mathrm{Hallucination} = \frac{\mathrm{FP}}{\mathrm{FP} + \mathrm{TN}} .
    \end{equation}
    This metric reflects the system’s tendency to incorrectly approve mismatched responses and return to the user, which is critical for compliance-sensitive applications.

    \item \textbf{Precision} evaluates the reliability of accepted responses and is defined as
    \begin{equation}
    \mathrm{Precision} = \frac{\mathrm{TP}}{\mathrm{TP} + \mathrm{FP}} .
    \end{equation}
    A higher precision indicates that accepted responses are more likely to be correct.

    \item \textbf{F1 Score} captures the balance between precision and recall and is defined as
    \begin{equation}
    \mathrm{F1} = \frac{2 \cdot \mathrm{Precision} \cdot \mathrm{Recall}}{\mathrm{Precision} + \mathrm{Recall}} ,
    \end{equation}
    where $\mathrm{Recall} = \frac{\mathrm{TP}}{\mathrm{TP} + \mathrm{FN}}$ measures the coverage of correctly accepted responses.
\end{itemize}

\subsection{Baseline Implementations}

In this section, we describe the implementation details of the LLM debate baseline and the neural ensemble model. 

\paragraph{Explainable LLM Debate.} We adopt a two-stage LLM-based debate framework, built on the method proposed in \cite{10.1145/3589334.3645381}, to enhance the process of LLM judgment.

In the first stage, the Qwen-2.5-7B model is prompted to act as a critical service agent and generate two single-sentence justifications explaining why a retrieved QA pair is and is not an appropriate response to the given user query. This stage focuses exclusively on identifying and explaining semantic misalignment, incomplete coverage, or contextual mismatch between the query and the retrieved response. The prompt is constrained to produce a single sentence without additional formatting to ensure consistency. Inference is performed in batches using deterministic decoding to reduce variance across samples.

In the second stage, the Llama-3.1-8B LLM serves as a final judge that integrates both positive and negative debate signals. Given the user query, the retrieved QA pair, and two justifications arguing for and against the relevance of the QA, the model outputs a binary decision indicating whether the retrieved QA fully resolves the user’s query. The decision is restricted to a Yes or No output, enforcing a clear acceptance or rejection outcome. This design encourages the model to reason over competing perspectives while maintaining a concise and unambiguous judgment.

\paragraph{Neural Ensemble Model.}
We implement a lightweight neural ensemble model to aggregate the decisions of multiple LLM judges in a query-dependent manner. For each query, the model takes a binary judgment vector $\mathbf{S} \in \{0,1\}^{K}$, where $s_i$ represents the judgment of the $i$-th LLM, and a dense query embedding $\mathbf{e}_q \in \mathbb{R}^{d}$ obtained from a pretrained sentence encoder. A two-layer feedforward gating network uses $\mathbf{e}_q$ to produce a query-dependent weight vector $\mathbf{g} \in (0,1)^{K}$, with ReLU activation in the hidden layer and sigmoid activation at the output. The element-wise product $\mathbf{S} \odot \mathbf{g}$ yields a reliability-adjusted vector, which is passed through a linear classifier and then through a sigmoid activation to produce the probability that the retrieved QA should be accepted.

The model is trained using binary cross-entropy loss with class weighting to account for label imbalance, upweighting positive samples to improve recall. Optimization is performed with AdamW in mini-batches over 50 epochs.

\subsection{Hyperparameters}

Hyperparameter tuning was performed to identify optimal configurations for the proposed latent ensemble model. The model uses a hidden dimension of 512, a dropout rate of 0.3, and a learning rate of 1e-3. Training employs 256 Monte Carlo samples with 10 variational inference iterations, while evaluation uses 1,024 Monte Carlo samples with 60 variational inference iterations. These settings were found to effectively balance model performance and training stability.

\begin{table*}[!t]
\renewcommand{\arraystretch}{1.2} 
\centering
\small
\resizebox{\linewidth}{!}{
\begin{tabular}{
>{\raggedright\arraybackslash}m{1.7cm} 
>{\raggedright\arraybackslash}m{6.8cm}
>{\raggedright\arraybackslash}m{1.2cm}
>{\raggedright\arraybackslash}m{7.2cm}
}
\toprule
\textbf{Dataset} & \textbf{Verified Response QA} & \textbf{Type} & \textbf{Query} \\
\midrule
\multirow{2}{*}{\textbf{CovidQA}} 
& \textbf{Q}: What type of \textbf{coronavirus} was detected in \textit{R. affinis} and \textit{R. sinicus} species?
& Matched & What \textbf{coronavirus} variant was discovered in the bat species \textit{R. affinis} and \textit{R. sinicus}? \\
\cline{3-4}
& \textbf{A}: A novel alpha-CoV, BtCoV/Rh/YN2012. & Unmatched & What viruses were found in \textit{Rhinolophus affinis} and \textit{Rhinolophus sinicus} species? \textcolor{red}{[General Query]} \\ \midrule
\multirow{2}{*}{\textbf{ExpertQA}} 
& \textbf{Q}: What are the requirements for claiming inheritance as per the Intestate Succession Act in \textbf{South Africa}?
& Matched & How can one claim inheritance under the Intestate Succession Act in \textbf{South Africa}? \\
\cline{3-4}
& \textbf{A}: The deceased must have been domiciled in South Africa at the time of death. & Unmatched & In order to collect an inheritance through intestacy, which stipulations from the Act apply? \textcolor{red}{[Ambiguous Entity]} \\ \midrule
\multirow{2}{*}{\textbf{HotpotQA}} 
& \textbf{Q}: The English chef known for his back-to-basics philosophy has a sister named what?
& Matched & What is the name of the sister of the English chef who advocates a back-to-basics approach? \\
\cline{3-4}
& \textbf{A}: Jane M. Fearnley-Whittingstall & Unmatched & Can you tell me the given name of this English chef's sister? \textcolor{red}{[Lack of Context]} \\ \midrule
\multirow{2}{*}{\textbf{HAGRID}} 
& \textbf{Q}: Who invented the first synthetic polymer?
& Matched & Who first engineered a polymer that was fully synthetic? \\
\cline{3-4}
& \textbf{A}: Leo Baekeland invented the first synthetic polymer in 1907 & Unmatched & What scientist is credited with making the very first synthetic material? \textcolor{red}{[Mismatched Entity]} \\ \midrule
\multirow{2}{*}{\textbf{MSMARCO}} 
& \textbf{Q}: What is the \textbf{average} annual salary of a dancer?
& Matched & On average, how much do dancers make in a year? \\
\cline{3-4}
& \textbf{A}: The average annual salary of a dancer is \$30,683. & Unmatched & What would you pay a typical \textbf{professional} dancer yearly? \textcolor{red}{[Mismatched Condition]} \\
\bottomrule
\end{tabular}
}
\caption{Examples of verified QA pairs from the five datasets, with two types of user query (matched and unmatched). The reasons for the unmatched queries are highlighted in red.}
\label{tab:positive_negative_query_example}
\end{table*}

\subsection{Prompt Tuning}
We employ PromptWizard with the \textbf{critique\_n\_refine} technique for automated prompt optimization \cite{agarwal-etal-2025-promptwizard}, using GPT-4o-mini as a cost-efficient model.

A dataset of 50 training and 30 test samples was manually prepared, with each sample consisting of a user query, the top-1 retrieved QA pair, and a binary ground-truth label (\texttt{true} if the QA perfectly matches the query intent, \texttt{false} otherwise), annotated by a human expert. The prompt-tuning process includes iteratively mutating and refining the base task description and instructions through critique, producing expert profiles, intent keywords, and reasoning chains. The optimized prompt specifies a meticulous QA evaluation role, clear task formulation, and a strict \texttt{Yes/No} output format. 

From preliminary evaluation, we observed that adding few-shot examples with reasoning distracts the model’s context, degrades performance, and significantly increases inference cost and latency. Therefore, the final prompt is zero-shot.

\begin{tcolorbox}[top=1pt, bottom=1pt, left=1pt, right=1pt]
  \textbf{Judgment Prompt - } You are a meticulous QA evaluation agent. Your task is to determine if a retrieved question-answer pair fully resolves the user’s query. Consider whether the QA pair addresses all parts of the query and maintains semantic equivalence. Think about how you could measure whether the response makes progress toward fully answering the user’s question. Respond only with 'Yes' or 'No', no explanation.
  
  \vspace{1em}
 User Query: \{user\_query\}
  
  \vspace{0.5em}
  Retrieved Question: \{candidate\_Q\}
  
  Retrieved Answer: \{candidate\_A\}

  \vspace{0.5em}
  Output 'Yes' if the retrieved QA is a perfect match, otherwise 'No'.
\end{tcolorbox}

\section{Data Annotation and Examples}

To ensure high-quality data, we recruited five human annotators fluent in English, each with at least one year of experience in data annotation or quality assurance. Prior to annotation, all annotators participated in a mandatory training session, during which the authors presented 50 query-response pairs with gold-standard matching judgments. This step ensured a shared understanding of labeling criteria and promoted consistent annotations.

Table \ref{tab:positive_negative_query_example} presents examples of QA pairs from five different datasets, where each QA is paired with two user queries: one that correctly matches the QA and one that represents a mismatch. These examples demonstrate the critical role of precise semantic judgment. Although all QAs are drawn from a verified knowledge base, returning an incorrect QA does not produce factually wrong information. However, the mismatched response fails to satisfy the user’s intent, potentially leading to confusion or dissatisfaction. This highlights the need for a careful assessment of query-response alignment at the intent level, which is essential for maintaining both the reliability of the system and a positive user experience.

\begin{figure}
    \centering
    \includegraphics[width=1\linewidth]{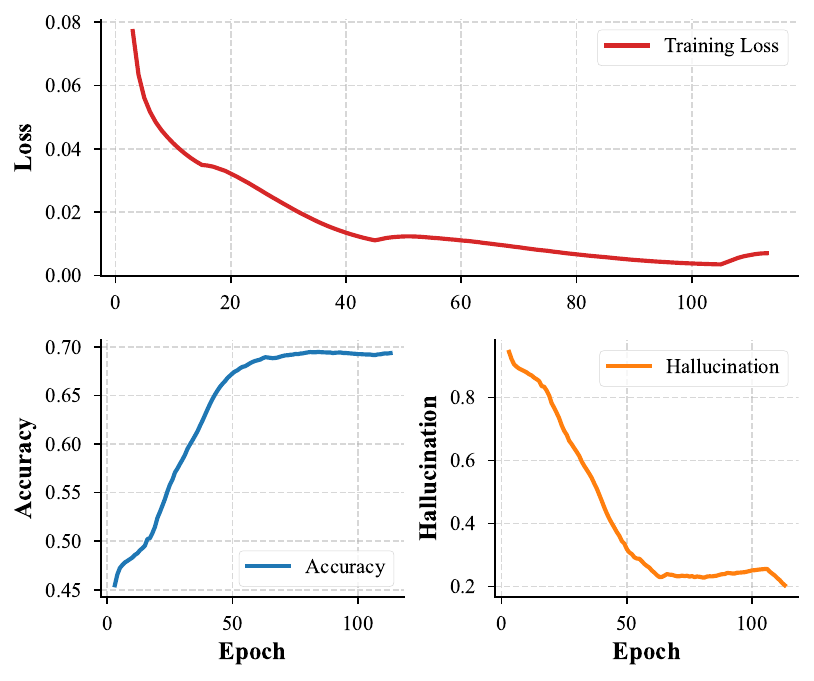}
    \vspace{-2em}
    \caption{Convergence behaviour of the proposed latent ensemble model.}
    \label{fig:convergence}
\end{figure}

\section{Model Convergence}
\label{appendix:convergence}

Figure \ref{fig:convergence} illustrates the stable convergence of the latent ensemble model over 100 epochs. Training loss initially decreases sharply from 0.08 to near 0.02, then plateaus smoothly, indicating effective optimization without overfitting. Accuracy rises rapidly from 0.45 to around 0.7, stabilizing thereafter, reflecting improved predictive performance. The hallucination rate declines steadily from 0.9 to below 0.2, demonstrating successful mitigation of unreliable outputs. These trends confirm the model’s robustness in jointly enhancing accuracy while reducing hallucinations, avoiding pitfalls such as excessive refusal observed in LLM debate and neural model baseline.

\section{Evaluation on Chinese Dataset}
\label{appendix:chinese}
To further evaluate the proposed method in a practical setting, we leverage a proprietary, industrial-grade dataset in Chinese, which captures richer linguistic diversity and domain-specific complexity\footnote{The data will be publicly released upon acceptance.}. This dataset is curated for conversational AI tasks in the financial services domain, covering a wide range of user inquiries related to insurance policies, claims, quotes, and general financial advice. It contains over 50,000 QA pairs verified by 14 financial experts, along with a corresponding unstructured knowledge base ($D_2$) of over 100,000 documents, including policy manuals, legal disclaimers, and product descriptions. For training, we synthetically generated 10,000 queries with a balanced distribution of positive and negative classes. For evaluation, we collected a test set of 2,000 realistic user queries, each labeled by human experts for ground-truth relevance with the top-1 retrieved candidate response from $D_1$.

To enhance Chinese language capability and reveal the effect of model selection, we replace LLaMA and Gemma with DeepSeek-v2 (13B parameters) and Baichuan-7B, both of which demonstrate stronger performance in Chinese comprehension and reasoning. Table \ref{tab:chinese correlation} shows the pairwise Pearson correlations among the five LLM judges on this dataset. The results reveal heterogeneous correlations across models, consistent with observations from the English dataset, confirming the presence of diverse dependencies among LLM judgments.

\begin{table}[!t]
\centering
\resizebox{\columnwidth}{!}{
\begin{tabular}{lccccc}
\toprule
\textbf{Model} & \textbf{ChatGLM} & \textbf{Baichuan} & \textbf{Qwen} & \textbf{Mistral} & \textbf{DeepSeek-v2} \\
\midrule
ChatGLM & 1.000 & 0.289 & 0.286 & 0.250 & 0.458 \\
Baichuan & 0.289 & 1.000 & 0.208 & 0.208 & 0.267 \\
Qwen & 0.286 & 0.208 & 1.000 & 0.394 & 0.360 \\
Mistral & 0.250 & 0.208 & 0.394 & 1.000 & 0.241 \\
DeepSeek-v2 & 0.458 & 0.267 & 0.360 & 0.241 & 1.000 \\
\bottomrule
\end{tabular}
}
\caption{Pairwise Pearson correlation of five LLM judges on the Chinese dataset.}
\label{tab:chinese correlation}
\end{table}

\begin{table}[!t]
\small
\centering
\resizebox{\columnwidth}{!}{
\begin{tabular}{lccc}
\toprule
\textbf{Model} & \textbf{Accuracy} & \textbf{Recall} & \textbf{F1-Score} \\
\midrule
BGE@0.85 & 0.83 & 0.80 & 0.82 \\
Single LLM & 0.78 & 0.75 & 0.72 \\
Majority Vote & 0.80 & 0.78 & 0.75 \\
\textbf{Latent Model (ours)} & \textbf{0.95} & \textbf{0.94} & \textbf{0.95} \\
\bottomrule
\end{tabular}}
\caption{Comparison of baseline methods and the proposed latent ensemble on the Chinese dataset.}
\label{tab:experiment_results_chinese}
\end{table}

Table \ref{tab:experiment_results_chinese} compares the performance of baseline methods against the proposed latent ensemble model. Unlike the English dataset, these Chinese queries do not strictly follow a generation schema, exhibiting higher linguistic diversity and slightly lower overall complexity, as some queries are loosely connected topically and can be distinguished more easily. While baseline methods achieve relatively high performance (around 80\% accuracy), the latent model consistently outperforms all baselines, achieving the highest accuracy, recall, and F1 Score. This demonstrates its strong ability to aggregate heterogeneous LLM judgments, correct errors from individual models, and robustly produce accurate relevance labels, highlighting that its superior performance is largely invariant to both model selection and language.

Overall, these additional results confirm that the query-adaptive latent ensemble is effective across languages and domains, handling heterogeneous LLM judgments and producing reliable outputs in retrieval-based service systems.

\section{Toy Example of Query-Adaptive Latent Ensemble}

To provide an intuitive understanding of the proposed Query-Adaptive Latent Ensemble, we present a toy example that illustrates the inference process. This example demonstrates how query-dependent latent competences are inferred, how heterogeneous LLM judgments are reconciled, and how the final relevance decision is produced. For simplicity, we consider a scenario with three LLM judges, using the following example from CovidQA for demonstration purposes.
        
\vspace{0.3em}
\noindent
\textbf{Query:} ``What type of coronavirus was detected in R. affinis and R. sinicus species?''

\vspace{0.3em}
\noindent
\textbf{Candidate QA:} ``What viruses were found in Rhinolophus affinis and Rhinolophus sinicus species?''

\vspace{0.3em}
\noindent
\textbf{Ground Truth:} 0 (unmatched response)

\paragraph{Query-Dependent Latent Competence.}

The model first encodes the query into a dense embedding $\mathbf{e}_q$. MuNet and SigmaNet then produce query-dependent latent means and variances for each judge. For our toy example, assume the latent means are:
\[
\boldsymbol{\mu}(\mathbf{e}_q) = (0.2, 0.3, 0.9),
\]
representing the query-dependent competence of Judges 1, 2, and 3, respectively. Let \(\mathbf{Z} = \boldsymbol{\mu}(\mathbf{e}_q)\) for this example, and let the observed LLM votes be:
\[
\mathbf{s} = (1, 1, 0).
\]

Judges 1 and 2 have low competence despite positive votes, whereas Judge 3 has high competence and votes correctly. In a simple majority-voting scheme, the final decision would be ``matched,'' which is inconsistent with the ground truth.

\paragraph{Compatibility and Interaction.}

The model then combines the latent competences with the observed votes and applies a gating mechanism to weight each judge’s contribution. Suppose the gate outputs are:
\[
\mathbf{g} = (0.3, 0.4, 0.9).
\]

Judges 1 and 2 are downweighted due to low competence, while Judge 3 dominates the interaction. The gated latent competences are combined through the interaction network to produce a consensus score:
\begin{equation*}
\begin{split}
\mathbf{Z} \odot \mathbf{g} &= (0.2 \cdot 0.3, 0.3 \cdot 0.4, 0.9 \cdot 0.9) \\
&= (0.06, 0.12, 0.81).
\end{split}
\end{equation*}
Passing this vector through the interaction network produces a consensus score:
\[
c = f_\text{int}(\mathbf{Z} \odot \mathbf{g}) \approx 0.9.
\]
This consensus score summarizes the weighted influence of all judges for the given query. It reflects both the observed votes and the query-dependent latent competences, modulated by the gating mechanism. A higher $c$ indicates stronger collective leaning toward a particular outcome and serves as input to the energy-based model, which integrates $c$ with other interaction potentials to compute the posterior probability \(P(y=1 \mid \mathbf{s}, \mathbf{e}_q)\).

\paragraph{Posterior Probability and Prediction.}

Finally, Monte Carlo samples of \(\mathbf{Z}\) are used to estimate the marginal probability of a positive match:
\[
P(y=1 \mid \mathbf{s}, \mathbf{e}_q) \approx 0.15.
\]

Applying a 0.5 threshold produces the final prediction:
\[
\hat{y} = 0.
\]

Despite two LLMs voting 1, the model correctly rejects the candidate QA, demonstrating robust hallucination mitigation.

\end{document}